\documentclass[sn-mathphys-num]{sn-jnl}

\usepackage{makecell}
\usepackage{graphicx}%
\usepackage{multirow}%
\usepackage{amsmath,amssymb,amsfonts}%
\usepackage{amsthm}%
\usepackage{mathrsfs}%
\usepackage[title]{appendix}%
\usepackage{xcolor}%
\usepackage{textcomp}%
\usepackage{manyfoot}%
\usepackage{booktabs}%
\usepackage{algorithm}%
\usepackage{algorithmicx}%
\usepackage{algpseudocode}%
\usepackage{listings}%
\usepackage{comment}
\usepackage{amssymb}
\usepackage{pgfplots}
\pgfplotsset{compat=1.18}
\usepackage{array}
\usepackage{tabularx}
\usepackage{tabulary}
\usepackage{pifont}
\usepackage{tikz}
\usepackage{threeparttable}
\usepackage{subcaption}
\usepackage{listings}
\usepackage{pifont}
\usepackage{xcolor} 
\usepackage{xcolor,pifont}
\newcommand*\colourcheck[1]{%
  \expandafter\newcommand\csname #1check\endcsname{\textcolor{#1}{\ding{52}}}%
}
\colourcheck{blue}
\colourcheck{green}
\colourcheck{red}

\newcommand*\colourcross[1]{%
  \expandafter\newcommand\csname #1cross\endcsname{\textcolor{#1}{\ding{55}}}%
}
\colourcross{blue}
\colourcross{green}
\colourcross{red}

\theoremstyle{thmstyleone}%
\theoremstyle{thmstyletwo}%

\theoremstyle{thmstylethree}%

\begin{document}

\title[Article Title]{Self-Contrastive Forward-Forward Algorithm}

\author*[1]{\fnm{Xing} \sur{Chen}}\email{xing.chen@cnrs-thales.fr}

\author[1]{\fnm{Dongshu} \sur{Liu}}\email{dongshu.liu@cnrs.fr}

\author[2,3]{\fnm{J\'er\'emie} \sur{Laydevant}}\email{jeremie.laydevant@gmail.com}

\author*[1]{\fnm{Julie} \sur{Grollier}}\email{julie.grollier@cnrs-thales.fr}

\affil*[1]{\orgdiv{Laboratoire Albert Fert}, \orgname{CNRS, Thales, Universit\'e Paris-Saclay}, \orgaddress{\street{1 av. A. Fresnel}, \city{Palaiseau}, \postcode{91767},  \country{France}}}

\affil[2]{\orgdiv{School of Applied and Engineering Physics}, \orgname{Cornell University}, \orgaddress{\city{ Ithaca}, \postcode{NY 14853}, \country{USA}}}

\affil[3]{\orgdiv{USRA Research}, \orgname{Institute for Advanced Computer Science}, \orgaddress{\city{Mountain View}, \postcode{CA 94035}, \country{USA}}}

\abstract{Agents that operate autonomously benefit from lifelong learning capabilities. However, compatible training algorithms must comply with the decentralized nature of these systems which imposes constraints on both the parameters counts and the computational resources. The Forward-Forward (FF) algorithm is one of these. FF relies only on feedforward operations, the same used for inference, for optimizing layer-wise objectives. This purely forward approach eliminates the need for transpose operations required in traditional backpropagation. 
Despite its potential, FF has failed to reach state-of-the-art performance on most standard benchmark tasks, in part due to unreliable negative data generation methods for unsupervised learning.

In this work, we propose Self-Contrastive Forward-Forward (SCFF) algorithm, a competitive training method aimed at closing this performance gap. Inspired by standard self-supervised contrastive learning for vision tasks, SCFF generates positive and negative inputs applicable across various datasets. The method demonstrates superior performance compared to existing unsupervised local learning algorithms on several benchmark datasets, including MNIST, CIFAR-10, STL-10 \textcolor{black}{ and Tiny ImageNet}.
We extends FF’s application to training recurrent neural networks, expanding its utility to sequential data tasks. These findings pave the way for high-accuracy, real-time learning on resource-constrained edge devices.}




\maketitle

\section{Introduction}\label{sec1}

On-chip edge learning is essential for adapting model weights in response to newly arrived, real-time, unlabeled personal data, making it especially suited for privacy-sensitive applications \cite{nahavandi2022application, cardinale2017wearable}. Unlike centralized cloud training, edge training also significantly reduces energy consumption, making it a more efficient and sustainable choice \cite{shi2016edge,khouas2024training}.
 
The surge in progress for inference using in-memory computing systems has yet to be matched in the realm of on-chip learning. Backpropagation \cite{richards2019deep}, the highest-accuracy algorithm for training neural networks, is particularly challenging to implement in hardware. Its non-locality requires storing all activation functions and their derivatives, and the backward pass necessitates bidirectional crossbar arrays, which doubles the number of transistors or the number of arrays, resulting in higher power and area consumption \cite{wang2024difficulties}. Local learning algorithms, which compute weight updates locally within each layer or neuron without relying on dependencies across the entire network, offer a promising solution to these challenges. Thus, there is a pressing need for local algorithms that are not only efficient for hardware implementations but can also handle time-series data and perform unsupervised learning, all while maintaining high accuracy \cite{markovic2020physics}. 

The upper section of Table \ref{tab:hardwarecompare} compares different categories of algorithms designed to address these challenges through their local updates, highlighting their specific advantages and issues for on-chip learning. While many algorithms effectively tackle specific aspects of on-chip learning, none achieve a comprehensive solution that simultaneously resolves all identified challenges.

\begin{table}
\centering
\tiny
\caption{Comparisons of the learning capabilities of different local learning methods for online training in crosssbar based hardware and their test accuracy [\%] on CIFAR-10, STL-10 \textcolor{black}{ and Tiny ImageNet} datasets. \textcolor{black}{For Tiny ImageNet, both the top-1 and top-5 accuracy are shown.} The symbols \greencheck, \redcross, and  - mean ``yes", ``no", and  ``no reported results" respectively. "Unidirectional inference" and "Unidirectional learning" means that weight matrices do not need to be transposed during inference or learning. "Unsupervised" means that the algorithm can handle unlabeled data. "Time series" 
refers to the ability of algorithlms to handle time series input with RNN architecture.}

\label{tab:hardwarecompare}
\begin{tabular}{@{}lcccccccccc@{}}
\toprule
\multirow{2}{*}{\makecell{\textbf{Algori} \\ \textbf{-thms}}}  & \multirow{2}{*}{\textbf{Local}}    & \multirow{2}{*}{\makecell{\textbf{Unidirec}
\\
\textbf{-tional} \\ 
\textbf{inference}}} & \multirow{2}{*}{\makecell{\textbf{Unidirec} 
\\
\textbf{-tional} \\ 
\textbf{learning}}} & \multirow{2}{*}{\makecell{\textbf{Simple} \\ \textbf{Activation }\\ \textbf{ function}}} &
\multirow{2}{*}{\makecell{\textbf{Unsuper} \\ \textbf{-vised}}} &
\multirow{2}{*}{\makecell{\textbf{Time} \\ \textbf{series}}} & \multirow{2}{*}{\makecell{\textbf{CIFAR} \\ \textbf{-10}}} &\multirow{2}{*}{\makecell{\textbf{STL} \\ \textbf{-10}}} &\multicolumn{2}{c}{\textbf{Tiny ImageNet}} \\  
\\ & & & & & & &  & & \textbf{Top-1} & \textbf{Top-5} \\
\midrule
   
Backprop             & \redcross & \greencheck             & \makecell{\redcross \\ (backward \\ pass)}      & \makecell{\greencheck \\ (ReLU)}   & \greencheck  & \greencheck    & 93.8 \cite{bandara2023guarding}                  & 91.70 \cite{bandara2023guarding}    
& \textcolor{black}{54.8} \cite{ozsoy2022self}\\
\hline
\makecell[l]{Energy \\ -based \\ }              & \greencheck           & \redcross             & \makecell{\redcross \\ (bidirectional\\ connections)}      & \makecell{\redcross \\ (HardSigmoid)}   & \redcross  & \redcross    & 92.3 \cite{hoier2023dual}                 & -      & -      \\
\hline
\makecell[l]{Hebb \\ -based \\ } & \greencheck   & \greencheck   & \makecell{\redcross \\ (Oja's weight\\ transpose)}     & \makecell{\redcross \\ (Softmax)}  & \greencheck  & \redcross        & {80.3 \cite{journe2022hebbian}}                 & {76.2 \cite{journe2022hebbian}} & -& {\textcolor{black}{37.0} \cite{lagani2021hebbian}}\\
\hline
\makecell[l]{Reward \\ Hebb \\ }           & \greencheck           & \greencheck             & \makecell{\redcross \\ (reward\\backward)}      & \makecell{\greencheck \\ (ReLU)}   & \greencheck  & \greencheck    & -                  & 73.6 \cite{illing2021local}    & -  \\ 
\hline
DFA
  & \greencheck& \greencheck               & \makecell{\redcross \\ (random\\backward)}     & \makecell{\greencheck \\ (Tanh)}  & \redcross  & \redcross        & 73.1 \cite{nokland2016direct}                & {-} & \textcolor{black}{32.1} \cite{webster2020learning} \\
\hline  
\\
 \multirow{5}{*}{FF}  & \multirow{5}{*}{\greencheck} & \multirow{5}{*}{\greencheck} & \greencheck & \multirow{5}{*}{\makecell{\greencheck \\ (ReLU)}} & \redcross & \multirow{5}{*}{\redcross} & 59.0 \cite{hinton2022forward} & - & {-} \\
 & & & \greencheck & & \redcross &  & 59.1 \cite{lee2023symba}& - & {-}\\
 & & & \greencheck & & \redcross &  & 78.1 \cite{papachristodoulou2024convolutional}& - & {-}\\
 & & & \redcross & & \redcross &  & 84.6 \cite{wu2024distance} & - & {-}\\
 & & & \greencheck & & \greencheck &  & 60.6 \cite{hwang2024employing} & - & {-} \\\\
\hline
\textbf{SCFF}     & \greencheck    & \greencheck               & \greencheck     & \makecell{\greencheck \\ (ReLU)}  & \greencheck  & \greencheck        & \textbf{80.8}                 & \textbf{77.3} & \textbf{\textcolor{black}{{35.7}}} & \textbf{\textcolor{black}{{59.8}}}\\
 \bottomrule
\end{tabular}

\end{table}

Direct Feedback alignment (DFA) offers an alternative to backpropagation by 
simplifying the learning process. Instead of computing exact gradients, it uses fixed, random feedback weights \cite{nokland2016direct,  frenkel2021learning}. This approach has been widely adopted in various hardware implementations for on-chip learning \cite{wang2024optical, launay2020hardware, filipovich2022silicon, neftci2017event}.
However, DFA still requires a backward pass through one or several synaptic arrays. Its performance also remains limited in deeper networks, particularly those integrating convolutional layers, despite impressive progress on other fronts \cite{launay2020direct}.

Hebbian-based learning methods \cite{hebb2005organization} present a strong interest for hardware implementations due to their simple two-factor learning rule that only depends on pre- and post- neuron activities. The recent development of algorithms such as SoftHebb \cite{journe2022hebbian} has considerably improved the accuracy of these approaches on unsupervised tasks. This success comes however at the cost of complex-to-implement softmax activation functions. Furthermore, bidirectional crossbar arrays are still needed for learning, as the weight update rule involves the transpose of the weight matrix. 

Reward-based Hebbian algorithms incorporate a third factor in the Hebbian learning rule in order to increase the accuracy \cite{illing2021local, bellec2020solution, halvagal2023combination}. The reward signal flows from the output to the intermediate layers through additional weight matrices, thus complicating the implementation.

Energy-based algorithms \cite{scellier2017equilibrium, stern2021supervised, hoier2023dual, scellier2024energy}, such as Equilibrium Propagation \cite{scellier2017equilibrium, laborieux2021scaling}, offer spatially local updates, simple activation functions and accuracies very close to BP. Those models have been implemented in specialized hardware, using architectures like network of variable resistors \cite{dillavou2022demonstration} and Ising machines \cite{laydevant2024training} to achieve efficient local updates and lower power consumption. However, they require hard-to-implement bidirectional connectivity as well as reaching network equilibrium for each input, which constitutes a hurdle for time-series classification. 

Therefore, further work is essential to develop learning algorithms that integrate all the key features necessary for practical implementation: local learning rules, unidirectional inference and learning, simple activation functions, and high accuracy in embedded application tasks, such as unsupervised learning of both static and dynamic data.

Recently, the Forward-Forward (FF) algorithm  \cite{hinton2022forward} emerged as a promising candidate for hardware-friendly learning due to its simplicity: it is local, unidirectional, works with simple activation functions like ReLU, and supports both static and sequential data. 
As a result, the FF algorithm has attracted significant interest since its introduction \cite{ororbia2023predictive, giampaolo2023investigating, zhao2023cascaded, lorberbom2024layer,tosato2023emergent,lee2023symba,papachristodoulou2024convolutional,reyes2024forward,scodellaro2023training,yang2023theory,gananath2024improved,hwang2024employing, kam2024memory}. Early demonstrations of FF in silico already cover a wide range of hardware platforms, including microcontrollers \cite{de2023mu}, in-memory computing systems utilizing Vertical NAND Flash Memory as synapses \cite{park2024chip}, and physical implementations where neurons are replaced by acoustic, optical, or microwave technologies \cite{momeni2023backpropagation,oguz2023forward}. However, its unsupervised learning performance has stagnated, particularly due to the challenge of generating high-quality negative data.

The lower section of Table \ref{tab:hardwarecompare} highlights recent advancements in the Forward-Forward (FF) algorithm \cite{hinton2022forward,papachristodoulou2024convolutional,lorberbom2024layer,hwang2024employing,lee2023symba,wu2024distance}, comparing its learning capabilities across various metrics. Notably, the accuracy of FF supervised training on CIFAR-10 dataset has improved significantly, rising from below 60\% \cite{hinton2022forward} to 84.7\% \cite{wu2024distance} by integrating random direct feedback connections, thereby closing the gap with leading supervised, purely local, forward algorithms \cite{kohan2023signal}. 

However, the performance of FF has plateaued in areas that are crucial for on-chip learning, such as unsupervised learning and time-series processing. In unsupervised learning, FF has shown only limited success \cite{hwang2024employing} and has struggled with datasets more challenging than MNIST \cite{deng2012mnist}, such as CIFAR-10 \cite{CIFAR-10}, \textcolor{black}{simplified ImageNet versions} \cite{le2015tiny} or the unlabeled dataset STL-10 \cite{coates2011analysis}. Additionally, FF is currently unable to effectively handle time-varying sequential data, which significantly limits its applicability to neuromorphic systems that often require processing dynamic, real-world inputs.

These challenges stem from the difficulty of generating ``negative" examples that are similar enough to the ``positive" training data to provide meaningful contrast, yet distinct enough for the network to learn effective representations. While creating such examples is relatively straightforward in supervised learning, where true or false labels can be integrated into the training process, there is currently no efficient, universal approach for generating positive and negative examples across all datasets. This limitation hinders FF's effectiveness in unsupervised learning and time-series processing, making it difficult to generalize across different data types and applications.

In this work, we propose the Self-Contrastive Forward-Forward (SCFF) method, which contrasts each data sample against itself to enable efficient learning. Inspired by self-supervised learning, SCFF generates positive and negative examples directly from the data, making FF applicable to unsupervised learning across a wide range of datasets. We demonstrate that SCFF not only extends FF’s capabilities to unsupervised tasks but also surpasses other local algorithms for unsupervised tasks on the MNIST, CIFAR-10, STL-10 \textcolor{black}{and Tiny Imagenet} datasets. Moreover, SCFF makes it possible to apply FF to sequential tasks, unlocking its potential for time-series data processing. Specifically, our contributions are as follows:

\begin{itemize}

 \item We propose the SCFF method, which efficiently generates positive and negative examples to train neural networks using the Forward-Forward algorithm in an unsupervised manner, applicable across a wide range of datasets. 

 \item We show that SCFF significantly outperforms existing unsupervised local learning algorithms in image classification tasks, achieving test accuracies of \textcolor{black}{98.70\% $\pm$ 0.01\%} on MNIST, \textcolor{black}{80.75\% $\pm$ 0.12\%} on CIFAR-10, and \textcolor{black}{77.30\% $\pm$ 0.12\%} on STL-10 (which includes a small number of labeled examples alongside a much larger set of unlabeled data). \textcolor{black}{ SCFF furthermore achieves a top-1 accuracy of {35.67\% $\pm$ 0.42\%} and a top-5
accuracy of 59.75\% $\pm$ 0.18\% on the more complex dataset Tiny ImageNet.}

  \item We present the first demonstration of the FF approach being successfully applied to sequential data. Our findings show that the proposed SCFF method effectively learns representations from time-varying sequential data using a recurrent neural network. In the Free Spoken Digit Dataset (FSDD), SCFF training improves accuracy by approximately 10 percentage points compared to reservoir computing methods.
  
  \item We conduct a theoretical and illustrative analysis of the distribution of negative examples within the data space. The analysis reveals that negative data points consistently position themselves between distinct clusters of positive examples, which suggests that negative examples play a crucial role in pushing apart and further separating adjacent positive clusters, thereby enhancing the efficiency of classification. 

\end{itemize}

Our results demonstrate the potential of Self-Contrastive Forward-Forward (SCFF) for efficient, layer-wise learning of meaningful representations in a local, purely forward, and unsupervised manner, enabling online training in hardware environments. In Section 2, we introduce SCFF, detailing its hardware applicability, innovative negative example generation method and its training procedure. Next, in section 3, we will present the results and discuss our findings. Finally, we will explore the relationship of SCFF to other purely forward and/or local learning methods in the discussion of section 4.

\begin{figure*}[ht]
\begin{center}
\includegraphics[width=\textwidth, clip=true, trim=2 2 2 2]{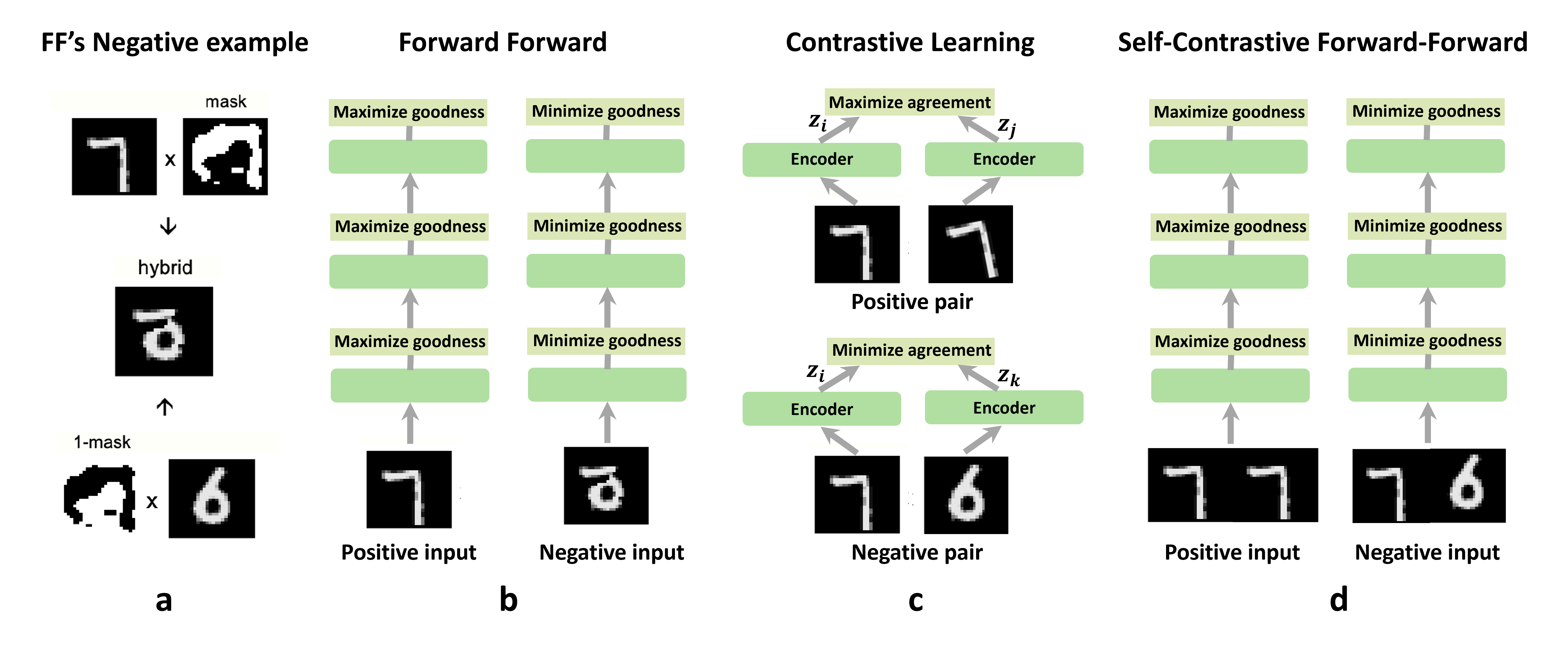}
\caption{\textbf{Comparative diagram illustrating three distinct unsupervised (self-supervised) learning paradigms.} \textbf{a}. Generation of a negative example is implemented by hybridization of two different images in the original FF paper \cite{hinton2022forward}. \textbf{b}. In Forward Forward (FF) Learning, the layer-wise loss function is defined so as to maximize the goodness for positive inputs (real images) and minimize the goodness for negative inputs, each of which is generated by corrupting the real image to form a fake image, as shown in \textbf{a}. \textbf{c}. In Contrastive Learning, the InfoNCE loss function determines the similarity between representations of two inputs (two different inputs or two same inputs but with different augmentations) in the end of the network \cite{chen2020simple}. \textbf{d}. Our proposed Contrastive Forward Forward Learning algorithm combines the principles of Forward Forward Learning and Contrastive Learning algorithms to maximize the goodness for concatenated similar pairs and minimize the goodness for dissimilar pairs with a layer-wise loss function.}

\label{comparison_algors}
\end{center}
\end{figure*}

\section{Self-Constrastive Forward-Forward algorithm}


The primary innovation of SCFF lies in its method for generating positive and negative examples, making it applicable to both supervised and unsupervised tasks across a wide variety of datasets. SCFF maintains the hardware-friendly, forward-only nature of FF while significantly enhancing its learning capabilities, particularly for unsupervised tasks and time-series classification.

\subsection{Creating the negative and positive examples}

In the FF algorithm, illustrated in Fig. \ref{comparison_algors}b, positive and negative examples are successively presented to the network input \cite{hinton2022forward}. A ``goodness'' score, related to local neuron activity, is computed at each layer, with the training objective being to maximize goodness for positive examples while minimizing it for negative ones. Therefore, the negative examples must be carefully crafted to effectively challenge the network, requiring more sophisticated approaches than basic noise injection or occlusion. For image data, the negative examples should maintain similar short-range correlations to the original data but differ significantly in long-range correlations to produce distinct shapes and categories. For the MNIST dataset, Hinton proposed generating negative examples by applying masks to different digit images and combining them to create hybrid images that look like digits but are not digits (Fig. \ref{comparison_algors}a). While this method works well for MNIST \cite{hinton2022forward}, it does not easily extend to more complex image databases like CIFAR-10, \textcolor{black}{ImageNet} and STL-10, or time series data, resulting in limited accuracy on these benchmarks. Therefore, further development of the FF algorithm must focus on devising robust methods for generating positive and negative examples that are applicable across diverse datasets.

As illustrated in Fig. \ref{comparison_algors}c, contrastive self-supervised learning methods share some similarity with FF. Negative and positive pairs are defined, where a positive pair consists of two different augmented views of the same data sample, and a negative pair of two different samples. Contrastive losses are employed to define the similarity in feature space between each image of a pair, with the training objective being to maximize similarity for positive pairs while minimizing it for negative ones.

\begin{figure*}[ht]
\vskip 0.2in
\begin{center}
\includegraphics[width=\textwidth, clip=true, trim=3 3 3 3]{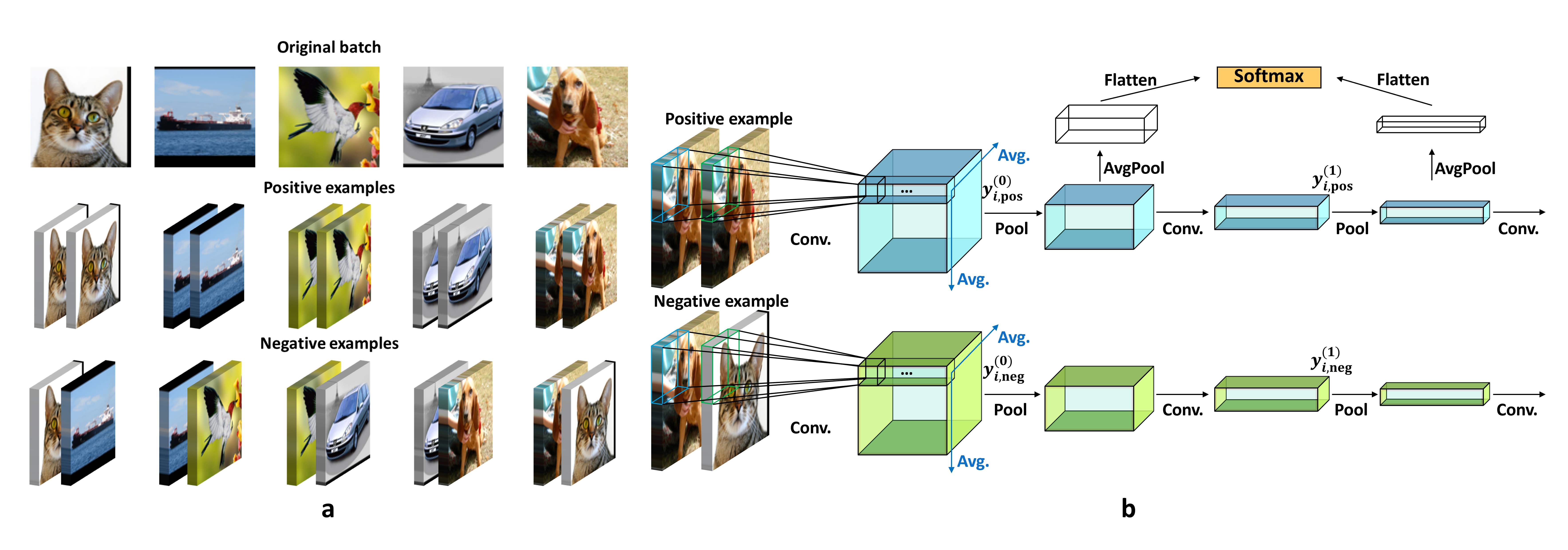}
\caption{\textbf{SCFF method for processing with Convolutional Neural Network Architecture}. \textbf{a}. The original batch of images (top row) is processed to generate positive (middle row) and negative examples (bottom row). \textbf{b}. The generated positive and negative examples undergo a series of convolutional (Conv.) and pooling (AvgPool or Maxpool) operations to extract relevant features. The output neurons which are extracted from each hidden layer after an external average pooling layer are then fed together into a softmax layer for final classification. }
\label{ContrastCNN}
\end{center}
\end{figure*}

In SCFF, we propose to directly take pairs of positive and negative images as inputs to the neural network (Fig. \ref{comparison_algors}d) instead of contrasting their representations in feature space as done in contrastive self-supervised learning (Fig. \ref{comparison_algors}c). More specifically, given a batch of $N$ training examples, and for a randomly selected example $\boldsymbol{x}_{k}$ ($k\in\{1, N\}$) in the batch, the positive example $\boldsymbol{x}_{i,\text{pos}}^{(0)}$ (the number 0 is the layer index) is the concatenation of two repeated $\boldsymbol{x}_{k}$, i.e., $\boldsymbol{x}_{i,\text{pos}}^{(0)} = [\boldsymbol{x}_{k}, \boldsymbol{x}_{k}]^T$. The negative example $\boldsymbol{x}_{j,\text{neg}}^{(0)}$ is obtained by concatenating $\boldsymbol{x}_{k}$  with another example $\boldsymbol{x}_{n}$ ($n\neq k$) in the batch, i.e., $\boldsymbol{x}_{j,\text{neg}}^{(0)} = [\boldsymbol{x}_{k}, \boldsymbol{x}_{n}]^T$ (or $[\boldsymbol{x}_{n}, \boldsymbol{x}_{k}]^T$ ). Fig. \ref{ContrastCNN}a shows some instances of generated positive and negative examples from the original training batch for the STL-10 dataset.

\textcolor{black}{Importantly, although the input size is doubled, the computational cost and memory usage of network training are not impacted compared to standard FF because the weight matrices connecting $\boldsymbol{x}_{k}$  and $\boldsymbol{x}_{n}$ can be identical. For example, in the fully connected network illustrated in Fig. \ref{comparison_algors}d, the outputs for the positive and negative examples from the first layer can be respectively written as:}

\begin{equation}
\textcolor{black}
{\boldsymbol{y}_{i,\text{pos}}^{(0)} = f(W_1\boldsymbol{x}_{k} + W_2\boldsymbol{x}_{k}), \quad \boldsymbol{y}_{j,\text{neg}}^{(0)} = f(W_1\boldsymbol{x}_{k} + W_2\boldsymbol{x}_{n}),}
\end{equation}
\textcolor{black}{where \( f \) is the ReLU activation function, and the weight matrices \( W_1 \) and \( W_2 \) correspond to the connections for each of the two images constituting the input. In practice, we set \( W_1 = W_2 \) because the gradient of the loss function with respect to \( W_1 \) and \( W_2 \) converges to the same value. This results in a reformulation where the inputs are summed rather than concatenated, yielding:}

\begin{equation}
\textcolor{black}{\boldsymbol{y}_{i,\text{pos}}^{(0)} = f(W(\boldsymbol{x}_{k} + \boldsymbol{x}_{k})), \quad \boldsymbol{y}_{j,\text{neg}}^{(0)} = f(W(\boldsymbol{x}_{k} + \boldsymbol{x}_{n})),}
\end{equation}
\textcolor{black}{where \( W = W_1 = W_2 \). This reformulation ensures that the effective input size remains unchanged compared to a standard single-image input, and the feature map dimensions in subsequent layers are not affected by the initial pairing operation. Unlike naive concatenation, which would increase feature map width, our approach maintains computational efficiency while preserving the relational structure of the input pairs. 
}

\textcolor{black}{Intuitively, this can be understood by recognizing that swapping the positions of \( \boldsymbol{x}_{k} \) and \( \boldsymbol{x}_{n} \) in the input should not affect the output neural activities. A rigorous mathematical proof of the convergence of matrices $W_1$ and $W_2$ is provided in Appendix A.}

\subsection{Training procedure}
The training procedure builds upon the FF framework, enhanced by additional unsupervised training techniques: 

\begin{itemize}
    \item \textbf{\textcolor{black}{Greedy} Layer-wise Training \textcolor{black}{/ Joint Training}:} 
    \textcolor{black}{SCFF supports two training paradigms: greedy layer-wise training and joint training, allowing for flexibility in optimization strategy based on computational constraints and architectural requirements. In greedy layer-wise training}, each layer of the network is fully trained before proceeding to the next layer. \textcolor{black}{In joint training, all layers are updated simultaneously using SCFF’s local learning rule in each iteration.} 
    After unsupervised training with SCFF, we froze the network and trained a linear downstream classifier \cite{alain2016understanding,journe2022hebbian} with the back-propagation method on representations created by the network using the labeled data. The linear classifier was optimized using cross-entropy loss. The accuracy of this classification serves as a measure of the quality of the learned representations.
    \item \textbf{Goodness Function:} SCFF employs a `goodness' function at each layer, similar to FF. The goodness score, $G_{i}^{(l)}= \frac{1}{M^{(l)}} \sum_m{y_{i,m}^{2(l)}}$, where $l$ is the layer index and $m$ represents the neuron index, calculated as the sum of squared activations, is optimized such that positive examples have higher goodness than negative examples. 
    \item \textbf{Loss Function:} Predefined positive and negative examples are successively presented to the network’s input. 
The possibility of a positive example $\boldsymbol{x}_{i}$ being recognized as positive and a negative example $\boldsymbol{x}_{j}$ being recognized as negative by the network are defined as $p_{\text{pos}}(\boldsymbol{x}_{i}) = \sigma(G_{i}^{(l)} - \Theta_{\text{pos}}^{(l)})$ and 
$p_{\text{neg}}(\boldsymbol{x}_{j}) = \sigma( \Theta_{\text{neg}}^{(l)} - G_{j}^{(l)})$ respectively. The sigmoid function $\sigma(x) = \frac{1}{1+e^{-x}}$ evaluates the effectiveness of the separation, where $\Theta_{\text{pos}}^{(l)}$ and $\Theta_{\text{neg}}^{(l)}$ are fixed values that serve as hyperparameters of the network. 

The loss function encourages the network to increase the goodness for positive examples so that it significantly exceeds the threshold ($p_{\text{pos}}(\boldsymbol{x}_{i}) \rightarrow$ 1) and to decrease the goodness score for negative input examples so that it falls well below the threshold ($p_{\text{neg}}(\boldsymbol{x}_{j}) \rightarrow$ 1). At layer $l$, the loss function is defined as:  

\begin{align}
    \mathcal{L}_{\text{FF}}
  &= -\mathbb{E}_{\boldsymbol{x}_{i}\sim \text{pos}}\text{log}p_{\text{pos}}(\boldsymbol{x}_{i}) - \mathbb{E}_{\boldsymbol{x}_{j}\sim \text{neg}}\text{log}p_{\text{neg}}(\boldsymbol{x}_{j})\nonumber\\
  &= -\mathbb{E}_{G_{i,\text{pos}}^{(l)}}\left[\text{log}\sigma(G_{i,\text{pos}}^{(l)}-\Theta_{\text{pos}}^{(l)}) \right]
-\mathbb{E}_{G_{j,\text{neg}}^{(l)}}\left[\text{log}\sigma( \Theta_{\text{neg}}^{(l)} -G_{j,\text{neg}}^{(l)} )\right]
\end{align}
where $G_{i,\text{pos}}^{(l)}$ and $G_{i,\text{neg}}^{(l)}$ respectively correspond to the goodness for the positive and negative examples input at layer $l$. The final loss is computed over all ${N}$ examples in the batch. 

\item \textbf{Normalization and Standardization}

To ensure stability during training, SCFF applies two key pre-processing steps: 
\begin{itemize}
    \item \textbf{Dataset-wide Normalization:} Each input image is normalized by subtracting the mean and dividing by the standard deviation, calculated per channel across the entire dataset. 
    \item \textbf{Per-image Standardization:} Each individual image is standardized so that its pixel values have a mean of 0 and a standard deviation of 1. This ensures consistent input scaling, which is particularly important for unsupervised learning. 
\end{itemize}
Other techniques include the ``Triangle" method \cite{coates2011analysis,miconi2021hebbian} for transmitting information between layers, adding a penalty term in the loss function to ensure stable training, and applying an extra pooling layer to retrieve information at each layer. For further details, see the Methods section. All details about the impact of hyperparameters and the training of the linear classifier are provided in Appendix \textcolor{black}{G} and \textcolor{black}{H}.

\end{itemize}

\section{Results}
 We evaluate SCFF on different image datasets including MNIST \cite{deng2012mnist}, CIFAR-10 \cite{CIFAR-10}, \textcolor{black}{Tiny ImageNet \cite{le2015tiny}} and STL-10 \cite{coates2011analysis} (results in Table \ref{tab:hardwarecompare} and Fig. \ref{fig:layer-acc}), as well as an audio dataset Free Spoken Digit Dataset (FSDD) \cite{jackson2018jakobovski} (results in Fig. \ref{fig:rnn}).  Across all benchmarks, SCFF surpasses all state-of-the-art algorithms that combine a local learning rule with fully-forward operation, while maintaining its advantage in hardware adaptability. 

\subsection{Multilayer Perceptron (MLP): MNIST }


\textcolor{black}{On the MNIST dataset, SCFF achieves a test accuracy of \textcolor{black}{98.70\% $\pm$ 0.01\%} when trained on a two-layer fully-connected network (MLP) with 2000 hidden neurons per layer, which is comparable to the performance achieved by backpropagation \cite{hinton2022forward}.
This surpasses previously published benchmarks on other biologically-inspired algorithms applied to MLPs, including 97.8\% in \cite{Moraitis_2022}, 98.42\% in \cite{srinivasan2023forward} (supervised training), and 96.6\% in \cite{zhou2022activation}. The full comparisons are presented in Table \ref{tab:compareSSL}, \ref{tab:compareForward}, and \ref{tab:compareEnergy}.
}

\subsection{Convolutional Neural Networks (CNN): \textcolor{black}{MNIST}, CIFAR-10 and  \textcolor{black}{Tiny ImageNet} }

The convolutional neural network (CNN) processes three-dimensional color images. The original images are concatenated along the channel dimension to form positive or negative inputs (see Fig. \ref{ContrastCNN}). The output of each convolutional layer is represented as a three-dimensional vector $\boldsymbol{y}_{i,\text{pos}}^{(l)}$ (or $\boldsymbol{y}_{i,\text{neg}}^{(l)}$) $\in \mathbb{R}^{C \times H\times W}$. The Loss function at layer $l$ is then defined as:

\begin{align}
    \mathcal{L}_{\text{SCFF}}
  =& -\mathbb{E}_{G_{i,\text{pos}}^{(l)}}\left[\frac{1}{H \times W}\sum_{h}^{H}\sum_{w}^{W}\text{log}\sigma( G_{i,h,w,\text{pos}}^{(l)} -\Theta_{\text{pos}}^{(l)})\right] \nonumber\\
&-\mathbb{E}_{G_{j,\text{neg}}^{(l)}}\left[\frac{1}{H \times W}\sum_{h}^{H}\sum_{w}^{W}\text{log}\sigma( \Theta_{\text{neg}}^{(l)} -G_{j,h,w,\text{neg}}^{(l)})\right] 
\end{align}
where the goodness of neural activities is calculated over the channels as $G_{i,h,w,\text{pos}}^{(l)} = \frac{1}{C}\sum_c{y_{i, c,h,w,\text{pos}}^{2(l)}}$ (or $G_{i,h,w,\text{neg}}^{(l)} = \frac{1}{C}\sum_c{y_{i, c,h,w,\text{neg}}^{2(l)}}$). 

\textcolor{black}{We evaluate SCFF using a three-layer CNN with the same architecture as the state-of-the-art SoftHebb, using 96, 384, and 1536 filters respectively \cite{journe2022hebbian}. On MNIST, our results show that SCFF achieves a test accuracy of 99.37\% $\pm$ 0.06\% when trained with a CNN, comparable to the 99.35\% achieved by SoftHebb \cite{journe2022hebbian}.}

\begin{figure}[!ht]
  \centering
  \begin{minipage}{0.45\textwidth}
    \centering
    \begin{tikzpicture}
      \begin{axis}[
        ybar,
        bar width=.35cm,
        enlarge x limits=0.22,
        width=1.1\textwidth,
        height=6cm,
        legend style={
          at={(1.3,1.15)},
          anchor=north,
          legend columns=-1,
          fill=none,
          draw=none,
          /tikz/every even column/.append style={column sep=0.5cm}
        },
        ylabel={Accuracy (\%)},
        symbolic x coords={1 layer,2 layers,3 layers},
        xtick=data,
        nodes near coords,
        every node near coord/.append style={font=\footnotesize, /pgf/number format/.cd, fixed, precision=1},
        nodes near coords align={vertical},
        ymin=50, ymax=90,
        title={\textbf{a} CIFAR-10},
        title style={at={(0.5,-0.2)},anchor=north},
        tick align=inside,
      ]
      \addplot+[area legend, bar shift=-0.25cm, error bars/.cd, y dir=both, y explicit] coordinates {(1 layer,72.05) +- (0, 0.31) (2 layers,78.28) +- (0, 0.38) (3 layers,80.75) +- (0, 0.12)};
      \addplot+[area legend, bar shift=0.25cm, error bars/.cd, y dir=both, y explicit] coordinates {(1 layer,74.48) +- (0, 0.06) (2 layers,84.06) +- (0, 0.12) (3 layers,84.92) +- (0, 0.2)};
      \legend{SCFF, Backprop}
      \end{axis}
    \end{tikzpicture}
  \end{minipage}%
  \hfill
  \begin{minipage}{0.5\textwidth}
    \centering
    \vspace{0.5cm}
    \begin{tikzpicture}
      \begin{axis}[
        ybar,
        bar width=.3cm,
        enlarge x limits=0.15,
        width=1.15\textwidth,
        height=6cm,
        legend style={
          at={(0.5,1.15)},
          anchor=north,
          legend columns=-1,
          fill=none,
          draw=none,
          /tikz/every even column/.append style={column sep=0.5cm}
        },
        symbolic x coords={1 layer,2 layers,3 layers,4 layers},
        xtick=data,
        nodes near coords,
        every node near coord/.append style={font=\footnotesize, /pgf/number format/.cd, fixed, precision=1},
        nodes near coords align={vertical},
        ymin=50, ymax=80,
        title={\textbf{b} STL-10},
        title style={at={(0.5,-0.2)},anchor=north},
        tick align=inside,
      ]
      \addplot+[area legend, bar shift=-0.25cm, error bars/.cd, y dir=both, y explicit] coordinates {(1 layer,66.15)  +- (0, 0.08) (2 layers, 72.35)  +- (0, 0.27) (3 layers,76)  +- (0, 0.26) (4 layers,77.3)  +- (0, 0.13)};
      \addplot+[area legend, bar shift=0.25cm, error bars/.cd, y dir=both, y explicit] coordinates {(1 layer,67.04)  +- (0, 0.38) (2 layers,73.287)  +- (0, 0.14) (3 layers,76.17)  +- (0, 0.30) (4 layers,77.025)  +- (0, 0.23)};
      \legend{ } 
      \end{axis}
    \end{tikzpicture}
  \end{minipage}
\caption{Comparison of test accuracy at different layers by using SCFF and Back-propagation methods on CIFAR-10 in \textbf{a} and on STL-10 dataset in \textbf{b}.}
\label{fig:layer-acc}
\end{figure}


For the CIFAR-10 dataset, 
SCFF achieves an accuracy of \textcolor{black}{80.75\% $\pm$ 0.12\%}, surpassing the previous state-of-the-art accuracies for purely-forward unsupervised learning, of 80.3\% on CIFAR-10 achieved using the SoftHebb algorithm \cite{journe2022hebbian}.

We also compared the test accuracies at each layer using SCFF and Backpropagation (BP) methods, as shown in Fig. \ref{fig:layer-acc}a. The comparison reveals that SCFF can effectively capture complex representations at deeper layers, similar to BP. Additionally, the evolution of accuracy with more layers indicates the scalability of SCFF and its robustness in deeper architectures.

\textcolor{black}{To further evaluate SCFF on more complex datasets, we apply an AlexNet-inspired architecture with five convolutional layers (filter sizes: 64-192-384-256-256) to Tiny ImageNet, which is a subset of ImageNet \cite{deng2009imagenet} containing 200 object classes. SCFF achieves a top-1 accuracy of {35.67\% $\pm$ 0.42\%} and a top-5 accuracy of {59.75\% $\pm$ 0.18\%}, significantly outperforming previous attempts to apply biologically inspired local learning algorithms to this database. Specifically, SCFF surpasses the Hebbian learning-based approach of Ref. \cite{lagani2021hebbian}, which achieved a top-5 accuracy of 36.99\%, and the Inference Learning Algorithm, which achieved a top-5 accuracy of 47.53\% \cite{alonso2024understanding}. These results highlight SCFF’s ability to learn hierarchical representations efficiently in larger-scale image classification tasks.}



\subsection{STL-10: Semi-Supervised Learning }

The STL-10 dataset is designed for semi-supervised learning tasks, where the majority of the data is unlabeled. We train SCFF using a four-layer convolutional neural network (CNN) with 96, 384, 1536  and 6144 filters respectively \cite{journe2022hebbian}. We compare its performance against both traditional supervised learning methods and other local learning algorithms. 

Notably, for STL-10, SCFF achieved a final layer 
performance of \textcolor{black}{77.30\% $\pm$ 0.12\%}, higher than the one of BP: 
\textcolor{black}{77.02\% $\pm$ 0.22\% }(Fig. \ref{fig:layer-acc}b). This is because the STL-10 dataset contains a large amount of unlabeled images, which limits the effectiveness of supervised BP training. By fine-tuning SCFF with end-to-end BP on the few labelled STL-10 examples, SCFF's
accuracy further improves to 80.13\%.
This demonstrates that SCFF is highly suitable for unsupervised pretraining followed by supervised BP training, making it ideal for semi-supervised learning approaches. 

Unlike other unsupervised learning methods, where the result is obtained solely from the final layer's output, SCFF integrates neuron information with the linear classifier from intermediate layers, leading to more comprehensive feature extraction \cite{hinton2022forward}. For CIFAR-10 (Fig. \ref{fig:layer-acc}a), the test accuracy for the two-layer and three-layer models 
was obtained by combining the outputs of all previous layers (pooled information for dimensionality reduction; see Methods section) before feeding them into the final linear classifier. For the STL-10 dataset, we combined the outputs from both the third and fourth layers for the final classification, resulting in a 1\% improvement in accuracy compared to using only the fourth layer's outputs as input to the linear classifier.

By visualizing and investigating the class activation map, which highlights the importance of each region of a feature map in relation to the model's output, we can intuitively observe that after four layers, more distinct and meaningful structures emerge (see Appendix \textcolor{black}{H}). Specifically, the activation maps corresponding to higher-layer features focus on the general contours and key objects within the input, facilitating improved feature extraction and classification \cite{lafabregue2021grad}.

\subsection{\textcolor{black}{Comparison of Greedy Layer-wise Training and Joint Training}}
\textcolor{black}{We compared the results of greedy layer-wise training and joint training on multiple datasets. Our results indicate that both approaches yield comparable performance. For instance, on CIFAR-10, joint training reaches an accuracy of {80.60\% $\pm$ 0.15\%}, whereas layer-wise training achieves 80.75\% $\pm$ 0.12\%. A similar trend is observed on STL-10, where joint training attains {77.14\% $\pm$ 0.04\%}, while layer-wise training achieves 77.30\% $\pm$ 0.14\%. These findings indicate that layer-wise training does not degrade feature representations, as confirmed by stable accuracy trends across deeper layers.}

\textcolor{black}{The similarity in performance between these two training strategies is due to the fact that weight updates in earlier layers do not affect the normalized inputs passed to subsequent layers \cite{hinton2022forward}. Specifically, all activations in a given layer scale by the same factor after an update, and layer normalization cancels out this scaling effect. This ensures that greedy layer-wise training prevents error accumulation, maintains stability, and yields feature representations comparable to joint training. Full derivations can be found in Appendix C.}

\textcolor{black}{Beyond accuracy, layer-wise training offers practical benefits: it enables more efficient hardware implementation and reduces memory constraints—making it particularly advantageous for neuromorphic computing. Additionally, training each layer independently allows for incremental learning and adaptability, making SCFF more flexible in scenarios where training resources are limited or where continual learning is desired.}

\subsection{Free Spoken Digit Dataset (FSDD): Sequential Data}

\textcolor{black}{Prior research in forward-only learning, including Fast Weights \cite{ba2016using}, differentiable plasticity \cite{miconi2018differentiable}, short-term plasticity \cite{rodriguez2022short, moraitis2020optimality}, and e-prop \cite{bellec2020solution}, has demonstrated the viability of forward-only approaches for sequence learning.} While the original FF paper \cite{hinton2022forward} includes a multi-layer recurrent neural network, it uses a static MNIST image repeated over time frames as input, with the objective of modeling top-down effects. Another implementation demonstrates a limited form of sequence learning with a fully connected network, but this architecture could not handle real-time sequential data due to the absence of recurrence. 
As a result, FF has yet to be extended to effectively handle recurrent network scenarios for time-varying inputs. One of SCFF’s most significant advancements over FF is its ability to handle time-series data. 

\begin{figure*}[ht]
\begin{center}
\includegraphics[width=1\textwidth, clip=true, trim=2 2 2 2]{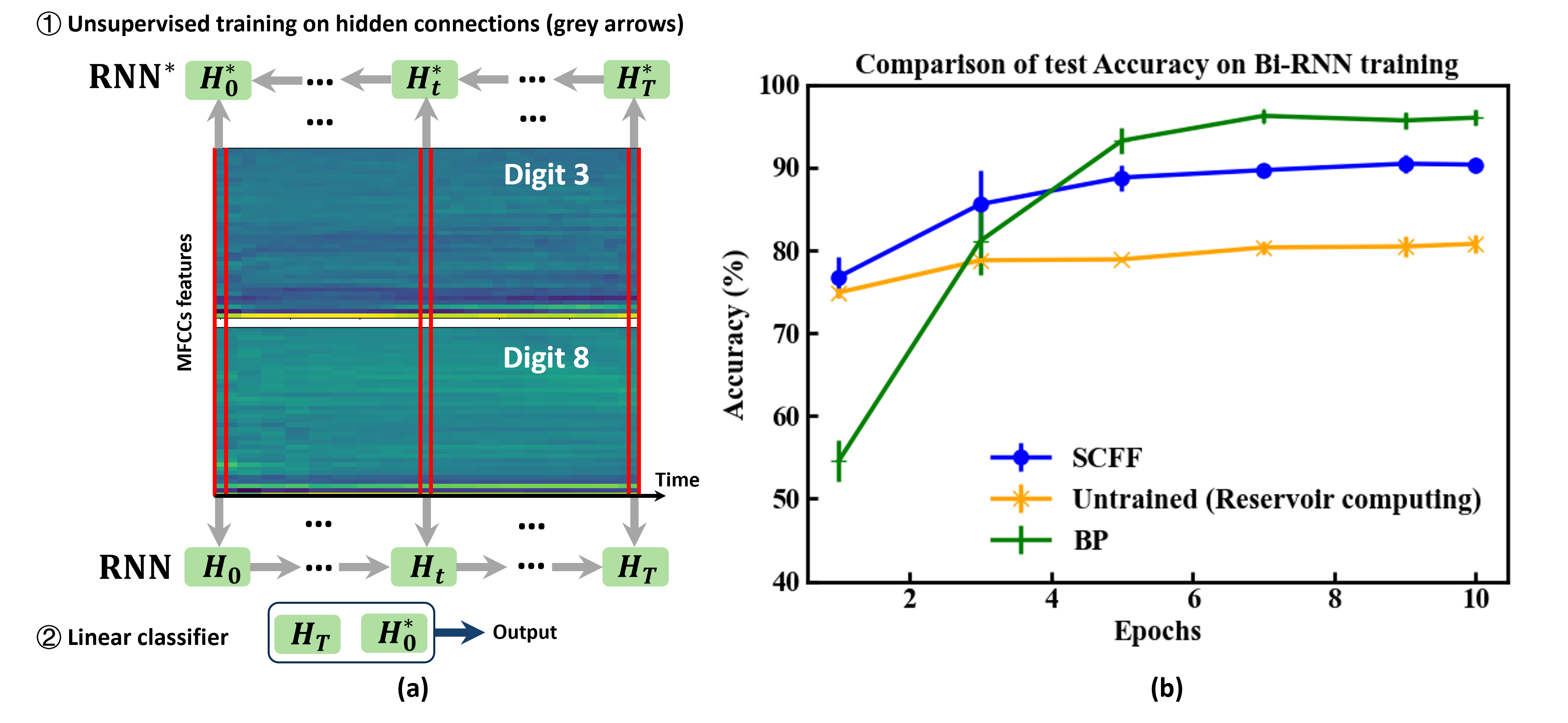}
\caption{Bi-directional RNN results on FSDD dataset. \textbf{a}. Training procedure of SCFF on a Bi-RNN. In the first stage, unsupervised training is performed on the hidden connections (both input-to-hidden and hidden-to-hidden transformations) using positive and negative examples. Positive examples are created by concatenating two identical MFCC feature vectors of a digit along the feature dimension, while negative examples are generated by concatenating MFCCs from two different digits, as illustrated in the figure. At each time step, the features are sequentially fed into the Bi-RNN (RNN and RNN$^*$). The red regions indicate features at different time steps. In the second stage, a linear classifier is trained using the final hidden states from both RNNs, i.e., $H_T$ and $H_0^*$ as inputs for classification task.
\textbf{b}. Comparison of test accuracy for the linear classifier trained on Bi-RNN outputs. The yellow curve represents accuracy with untrained (random) hidden neuron connections, the blue curve shows results from training with SCFF, the green curve shows Backprop results.}
\label{fig:rnn}
\end{center}
\end{figure*}

We employ the Free Spoken Digit Dataset (FSDD), a standard benchmark task for evaluating RNN training performance \cite{van2023nilrnn,limbacher2022memory, gokul2024poscuda}. The FSDD is a collection of audio recordings where speakers pronounce digits from 0 to 9 in English.
We follow the standard procedure that consist in extracting frequency domain information at different time intervals, here through Mel-Frequency Cepstral Coefficient (MFCC) features with 39 channels \cite{tiwari2010mfcc}. Plots of the evolution of MFCC features with time are shown in Fig. \ref{fig:rnn} for the digits 3 and 8. The SCFF method forms positive and negative examples by concatenating the same input for positive examples, and different ones for negative examples. Fig. \ref{fig:rnn}a presents a negative example which is generated by concatenating MFCC features from two different digits. The goal of the task is to recognize the digit after feeding in the full sequence, from the output of the network at the last time step. 

We train a Bi-directional Recurrent Neural Network (Bi-RNN) in an unsupervised way using the SCFF method to classify the digits. The procedure that we use for this purpose is illustrated in Fig. \ref{fig:rnn}a. Unlike conventional uni-directional RNNs, where the sequential input is processed step by step in a single direction, resulting in a sequence of hidden states from $H_0$ to $H_T$ (as depicted in the bottom RNN in Fig. \ref{fig:rnn}a), the Bi-RNN comprises two RNNs that process the input in parallel in both forward and backward directions. This results in hidden states evolving from $H_0$ to $H_T$ in the forward RNN and from $H_T^*$ to $H_0^*$ in the backward RNN$^*$, as shown in the top portion of the figure. The red regions in the figure highlight the features at different time steps. This bidirectional structure is particularly advantageous for tasks where context from both preceding and succeeding audio frames is critical, such as speech recognition, enhancing model performance compared to conventional uni-directional RNNs \cite{schuster1997bidirectional}.

The output of each directional RNN for a positive or negative input example is a two-dimentional vector $\boldsymbol{h}_{i}\in \mathbb{R}^{M \times T}$, where $T$ represents the number of time steps and $M$ denotes the number of hidden neurons. The loss function at layer $l$ is then defined as: 

\begin{align}
    \mathcal{L}_{\text{SCFF}}
  =& -\mathbb{E}_{G_{i,\text{pos}}^{(l)}}\left[\frac{1}{ T }\sum_{t}^{T}\text{log}\sigma( G_{i,t,\text{pos}}^{(l)}-\Theta_{\text{pos}}^{(l)})\right] \nonumber\\
&-\mathbb{E}_{G_{j,\text{neg}}^{(l)}}\left[\frac{1}{ T }\sum_{t}^{T}\text{log}\sigma(\Theta_{\text{neg}}^{(l)} -G_{j,t,\text{neg}}^{(l)})\right]
\end{align}
where the goodness of neural activities is calculated at each time step as $G_{i,t,\text{pos}}^{(l)} = \frac{1}{M}\sum_m{h_{i,t,m,\text{pos}}^{2(l)}}$ (or $G_{i,t,\text{neg}}^{(l)} = \frac{1}{M}\sum_m{h_{i,t,m,\text{neg}}^{2(l)}}$).

After the first stage of unsupervised training, a linear classifier is trained on the hidden states from the final time step in both directions, as shown in the bottom of the Fig. \ref{fig:rnn}a. The blue, orange and green curves in Fig. \ref{fig:rnn}b depict the test accuracy of the linear output classifier with hidden connections trained using SCFF, with random (untrained) hidden connections, and with Backpropagation methods, respectively. 

SCFF achieves a test accuracy of \textcolor{black}{90.33\% $\pm$ 0.94\% } when trained with a one-layer Bi-RNN containing 500 hidden neurons in each direction (refer to Appendix \textcolor{black}{F} for further architectural details). It is below the performance of BackPropagation Though Time that easily reaches \textcolor{black}{96.00\% $\pm$ 0.94\% } accuracy on this small task. However, SCFF avoids the issues of vanishing and exploding gradients of BPTT, as the gradients at each time step are calculated independently. This eliminates the dependency between time steps, providing a more stable training process which could be useful for future experiments on larger networks. 

Furthermore, the SCFF results are  well above the model with untrained (random) input and hidden connections which plateaus at \textcolor{black}{80.78\% $\pm$ 1.03\% }. Such model, in which only the output connections are trained, is akin to Reservoir Computing, a method that is often used to leverage physical systems on sequential data for neuromorphic applications \cite{abreu2020role}. SCFF provides a way to train these input and hidden layer connections in a simple, hardware-compatible way, and opens the path to a considerable gain of accuracy. This achievement opens the door for its extension to more complex tasks involving temporal sequences and its potential use in neuromorphic computing domains, such as dynamic vision sensors \cite{he2020comparing}.

Overall, this result constitutes the first successful application of the FF approach to sequential data in an unsupervised manner. \textcolor{black}{Future work could explore feedback mechanisms to enable top-down contextual modulation or extending SCFF with memory-augmented mechanisms, similar to Fast Weights \cite{ba2016using}, to improve temporal information retention.}


\section{Discussion}
\label{secDiscussion}

\subsection{Comparison to the original FF algorithm}


In SCFF, we have expanded the applicability of FF to complex unsupervised tasks beyond the MNIST dataset. The SCFF method achieves state-of-the-art (SOTA) accuracy for local methods on challenging datasets such as CIFAR-10, \textcolor{black}{Tiny ImageNet} and STL-10, largely outperforming the original FF algorithm (see Table \ref{tab:hardwarecompare}). This is a significant advancement, as it demonstrates that the method performs comparably to other local and forward-only algorithms in complex visual tasks, thereby broadening the scope and utility of FF to unlabeled data processing.

We have also shown that FF can solve sequential tasks. This extension is crucial for applications in time-series analysis and other domains where data is inherently sequential. By incorporating these improvements, our SCFF method not only overcomes the original limitations of FF but also sets a new benchmark for unsupervised learning algorithms in terms of versatility and performance.

\subsection{Analysis of the negative examples}

\textcolor{black}{The improvements brought by SCFF rely on a novel approach for consistently constructing positive and negative examples, in a way that can be applied to any database.}

 \begin{figure*}[ht]
\begin{center}
\includegraphics[width=1\textwidth, clip=true, trim=2 2 2 2]{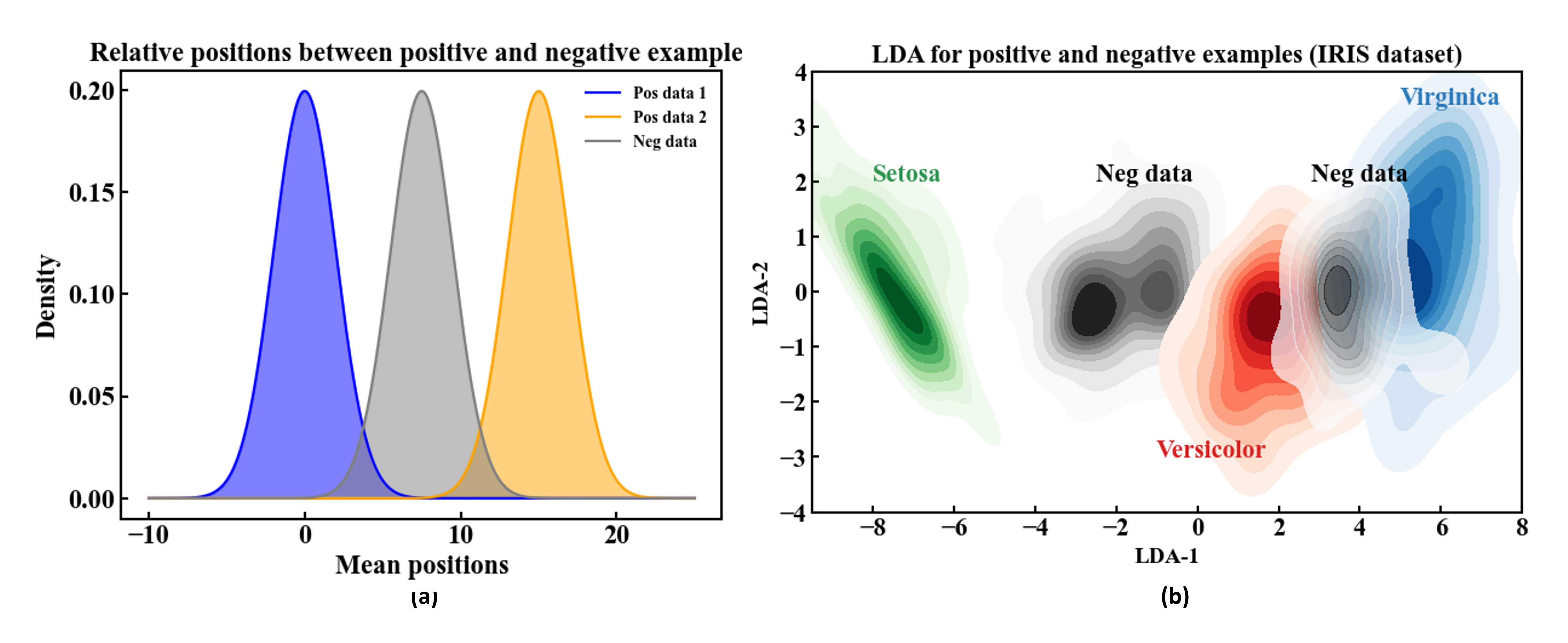}
\caption{Probability distributions of relative positions between positive and negative examples. \textbf{a} Theoretical distributions of positive examples from two different classes with distinct means ($2\mu_1 = 0$ and $2\mu_2 = 15$) and identical variance ($2\Sigma = 4$) are shown with blue and orange curves, respectively. The theoretical distribution of negative examples derived from the two classes using the formula \ref{eq:dis} is depicted by the grey curve. \textbf{b} Continuous probability density of LDA applied to the IRIS dataset, displaying contours for positive examples in green, red, and blue, and for negative examples in grey.}
\label{fig:prob_dist}
\end{center}
\end{figure*}

The effectiveness of the negative examples in SCFF can be understood through the lens of Noise Contrastive Estimation (NCE). In NCE, a key insight is that ``the noise distribution should be close to the data distribution, because otherwise, the classification problem might be too easy and would not require the system to learn much about the structure of the data" \cite{gutmann2010noise}. Our method of generating the positive and negative examples aligns with this principle if we treat the negative examples as ``noise data". We assume that the data samples for each class follow a multivariate Gaussian distribution with a shared covariance matrix $\Sigma$ and that each class is statistically independent of the others—assumptions commonly employed in various statistical models \cite{lee2019robust}. 

\textcolor{black}{
Since the input weight matrices are identical, i.e., \( W = W_1 = W_2 \), the input transformation simplifies to:  
\(\boldsymbol{y}_{i,\text{pos}}^{(0)} = f(W(\boldsymbol{x}_{k} + \boldsymbol{x}_{k})) = f(W(2\boldsymbol{x}_{k}))\),  
\(\boldsymbol{y}_{j,\text{neg}}^{(0)} = f(W(\boldsymbol{x}_{k} + \boldsymbol{x}_{n})).\)  
Thus, the positive and negative examples can be interpreted as:
}

\begin{equation}
\textcolor{black}{\boldsymbol{x}_{i,\text{pos}}^{(0)} = 2\boldsymbol{x}_{k}, \quad \boldsymbol{x}_{j,\text{neg}}^{(0)} = \boldsymbol{x}_{k} + \boldsymbol{x}_{n}.}
\end{equation}
Therefore, the distributions of positive examples $\boldsymbol{x}_{i,\text{pos}}^{(0)}$ and negative examples $\boldsymbol{x}_{j,\text{neg}}^{(0)}$ follow:
\begin{align}
\label{eq:dis}
\boldsymbol{x}_{i,\text{pos}}^{(0)}&\sim \mathcal{N}(2\boldsymbol{\mu}_1, 2\boldsymbol{\Sigma}) \nonumber\\
\boldsymbol{x}_{j,\text{neg}}^{(0)}&\sim \mathcal{N}(\boldsymbol{\mu}_1+\boldsymbol{\mu}_2, 2\boldsymbol{\Sigma})
\end{align}
where $\boldsymbol{\mu}_1$ and $\boldsymbol{\mu}_2$ are means of two different classes respectively.

Theoretically, the negative examples always lie somewhere between two different clusters of positive examples in the sample space, as illustrated in Fig. \ref{fig:prob_dist}a for the one-dimensional case. \textcolor{black}{Unlike conventional contrastive learning methods that require explicit hard negative sampling, SCFF inherently constructs negative examples that are statistically distinct yet sufficiently similar to the data distribution. This ensures that SCFF maintains a {strong contrastive signal} without relying on additional augmentation strategies.} For practical analysis with a real-world dataset, we visualized the distributions of positive and negative examples from the IRIS dataset \cite{misc_iris_53} using 2D linear discriminant analysis (LDA), which distinguishes between three different types of irises, as shown in Fig.\ref{fig:prob_dist}b. This visualization shows that the negative examples are positioned between different clusters of positive examples, suggesting that they contribute to pushing apart and further separating adjacent positive examples as they are mapped into higher-dimensional space during training. Additionally, negative examples are formed by combining two examples from different classes, enriching the diversity of negative examples and leading to more robust training. For a detailed analysis of how the LDA components evolve during training as the input data is mapped into the feature space and more theoretical results, please refer to Appendix B.

\subsection{Comparison to SOTA self-supervised learning (SSL)}

The SCFF method is inspired from self-supervised contrastive learning techniques \cite{chen2020simple, he2020momentum}. \textcolor{black}{While its purely local, layer-wise learning may limit its accuracy compared to end-to-end trained SSL models (Table \ref{tab:compareSSL}), this design offers unique advantages.} SCFF indeed operates without the auxiliary heads (multi-layer nonlinear projector) nor the complex regularization techniques required in global SSL methods, which simplifies its implementation and makes it more suitable for neuromorphic computing applications. \textcolor{black}{SCFF’s local loss and layer normalization also ensure stable activations and mitigate the vanishing gradient problems that typically arise in deep networks trained via backpropagation.}


\textcolor{black}{Another key distinction between SCFF and standard contrastive learning approaches lies in the way how negative samples are handled. Traditional  methods such as SimCLR \cite{chen2020simple} and MoCo \cite{he2020momentum} require large batches or memory banks to construct multiple negative pairs, often incorporating hard negatives and strong augmentations to enhance contrastive separation. These operations are computationally expensive and time-consuming. In contrast, SCFF maintains a fixed 1:1 ratio of positive to negative samples by optimizing a goodness-based objective, eliminating the need for additional transformations or augmentations. This formulation ensures that SCFF remains computationally efficient while still preserving a meaningful contrastive signal.}
\textcolor{black}{Future work could explore enhancements such as integrating top-down feedback connections and incorporating feedback-informed negative selection, where model predictions dynamically refine the selection of negative samples to enhance contrastive separation to further scale SCFF to modern deep networks like ResNet-50 and Vision Transformers (ViTs).} 




Recent developments in local versions of contrastive self-supervised learning have shown promising results \cite{laydevant2023benefits, tang2022biologically}. For instance, Laydevant et al. \cite{laydevant2023benefits} empirically demonstrated that layer-wise SSL objectives can be optimized rather than a single global one, achieving performance comparable to global optimization on datasets such as MNIST and CIFAR-10 (see Table \ref{tab:compareSSL}). However, this accuracy comes from multi-layer MLPs as projector heads at each layer, increasing the computational complexity. Illing et al. \cite{illing2021local} have shown that local plasticity rules, when applied through the CLAPP model, can successfully build deep hierarchical representations without the need for backpropagation. However, this method introduces additional processing along the time axis, which may add complexity when dealing with data that lacks temporal dynamics.


\begin{table}
\centering
\footnotesize
\caption{Test accuracy [\%] \textcolor{black}{(top-1)} comparison of SCFF with self-supervised Learning methods on MNIST, CIFAR-10, STL-10, \textcolor{black}{and Tiny ImageNet} datasets. The symbols \greencheck, \redcross, and  - mean ``yes", ``no", and  ``no reported results" respectively.}
\label{tab:compareSSL}
\begin{tabular}{@{}lccccccc@{}}
\toprule
\textbf{Method}  & \makecell{\textbf{Unsuper} \\ \textbf{-vised}}   & \textbf{MNIST} & \makecell{\textbf{{CIAFR}} \\ \textbf{{-10}}} & \makecell{\textbf{{STL}} \\ \textbf{{-10}}}   & \makecell{\textbf{\textcolor{black}{Tiny}} \\ \textbf{\textcolor{black}{-ImageNet}}} \\ \midrule
SimCLR (Chen et. al. 2020 \cite{chen2020simple})                     & \greencheck      & -       & 94.0                  & 89.7  & \textcolor{black}{53.4} \cite{robinson2020contrastive}      \\
Bio-SSL (Tang et. al. 2022 \cite{tang2022biologically})              & \greencheck      & -                            & 72.7      & 68.8   & -  \\
PNN-SSL (Laydevant et. al. 2023 \cite{laydevant2023benefits})                 & \redcross            &  96.6                & 77.0      & -   & -  \\
CLAPP (Illing et. al. 2021 \cite{illing2021local})                  & \greencheck  & -                            & -      & 73.6   & -  \\
\textbf{SCFF (ours) }                        & \greencheck             & \textbf{98.7}             & \textbf{80.8}                 & \textbf{77.3} & \textbf{\textcolor{black}{35.7}}\\
 \bottomrule
\end{tabular}

\end{table}

\subsection{Comparison to other forward-only methods}

The development of purely forward learning techniques has been historically driven by their potential for biologically plausible and neuromorphic computing applications \cite{srinivasan2023forward, zhou2022activation}. We compare recent and SOTA results of purely-forward methods with local learning rules to SCFF in Table \ref{tab:compareForward}.

Similar to Forward-Forward (FF), Pepita \cite{srinivasan2023forward} processes data samples in two forward passes. The first pass is identical to FF, while the input of the second pass is modulated by incorporating information about the error from the first forward pass through top-down feedback. Activation Learning \cite{zhou2022activation} builds on Hebb’s well-known proposal, discovering unsupervised features through local competitions among neurons. However, these methods do not yet scale to more complex tasks, limiting their potential applications.

\begin{table}
\centering
\footnotesize
\caption{Test accuracy [\%] comparison of SCFF with other forward-only methods on MNIST \textcolor{black}{(top-1)}, CIFAR-10 \textcolor{black}{(top-1)}, STL-10 \textcolor{black}{(top-1)} datasets, \textcolor{black}{and Tiny ImageNet (top-5)}. The symbols \greencheck, \redcross, and  - mean ``yes", ``no", and  ``no reported results" respectively.}
\label{tab:compareForward}
\begin{tabular}{@{}lccccccc@{}}
\toprule
\textbf{Method}  & \makecell{\textbf{Unsuper} \\ \textbf{-vised}}   & \textbf{MNIST}  & \makecell{\textbf{{CIAFR}} \\ \textbf{{-10}}} & \makecell{\textbf{{STL}} \\ \textbf{{-10}}} & \makecell{\textcolor{black}{\textbf{Tiny}} \\ \textcolor{black}{\textbf{ImageNet}} }  \\ \midrule
SigProp (Kohan et. al. 2023 \cite{kohan2023signal})                   & \redcross    & 98.2                           & 91.6         & -  & -  \\
PEPITA (Srinivasan et. al. 2023 \cite{srinivasan2023forward})                 & \redcross                & 98.4                & 53.8                & -  & -\\
Act. Learning (Zhou et. al. 2022 \cite{zhou2022activation})                 & \greencheck       & 97.1             & 58.7                & - & -\\
HardHebb (Miconi et. al. 2021 \cite{miconi2021hebbian})                 & \greencheck    & -             & 64.8                & -        & -        \\
HardHebb (Lagani et. al. 2021 \cite{lagani2021hebbian})                  & \greencheck      & 98.5 \cite{lagani2022comparing}             & 65.9         & -        & \textcolor{black}{37.0}               \\
Hebb-CHU (Krotov et. al. 2019 \cite{krotov2019unsupervised})               & \greencheck         & 98.5           & 50.8                & -             & -
\\
Hebb-PNorm (Grinberg et. al. 2019 \cite{grinberg2019local})                  & \greencheck     & -            & 72.2                & -            & -    
\\
SoftHebb (Journ{\'e} et. al. 2022 \cite{journe2022hebbian})                      & \greencheck            & 97.8             & 80.3                & 76.2 & -\\

\textbf{SCFF (ours) }                        & \greencheck             & \textbf{98.7}             & \textbf{80.8}                 & \textbf{77.3} & \textbf{\textcolor{black}{59.8}}\\
 \bottomrule
\end{tabular}
\end{table}

\begin{table}
\centering
\footnotesize
\caption{\textcolor{black}{Single-layer training performance (in terms of epoch duration) for different publicly available forward-only and local learning repositories, compared to SCFF,
on CIFAR-10. Accuracy results of the overall models from the respective papers on
CIFAR-10 are also reported. All experiments were conducted on a single NVIDIA RTX 4090 GPU.}}
\label{tab:trainingtime}
{\color{blue}
\begin{tabular}{@{}lccccc@{}}
\toprule
\textbf{Method} & \textbf{Unsupervised} & \textbf{Epoch duration (s) }    & \textbf{Acc. (\%)} \\ \midrule
SigProp (Kohan et. al. 2023 \cite{kohan2023signal})               &\redcross    & 7.5    & 91.6 \\
HardHebb (Miconi et. al. 2021 \cite{miconi2021hebbian})         & \greencheck        &  1.6    & 64.8           \\
HardHebb (Lagani et. al. 2022 \cite{lagani2022comparing})          & \greencheck        &  5.1     & 64.6             \\
SoftHebb (Journ{\'e} et. al. 2022 \cite{journe2022hebbian})         & \greencheck             & {2.6}             & 80.3               \\

\textbf{SCFF (ours) }             & \greencheck           & \textbf{1.2}                    & \textbf{80.8}      \\
 \bottomrule
\end{tabular}
}
\end{table}

Hebbian deep learning has also achieved remarkable progress recently \cite{miconi2021hebbian, lagani2022comparing, lagani2022fasthebb, journe2022hebbian}. These methods are purely local in space and can be applied purely locally in time, offering a biologically plausible approach to learning. Miconi \cite{miconi2021hebbian} demonstrated that Hebbian learning in hierarchical convolutional neural networks can be implemented with modern deep learning frameworks by using specific losses whose gradients produce the desired Hebbian updates. However, adding layers has not resulted in significant performance improvements on standard benchmarks \cite{miconi2021hebbian, lagani2022comparing}. Journe et al. \cite{journe2022hebbian} proposed using a simple softmax to implement a soft Winner-Takes-All (WTA) and derived a Hebbian-like plasticity rule (SoftHebb). With techniques like triangle activation and adjustable rectified polynomial units, SoftHebb achieves increased efficiency and biological compatibility, enhancing accuracy compared to state-of-the-art biologically plausible learning methods.

Our SCFF method brings the FF approach to accuracy levels comparable to SoftHebb, effectively bridging the gap between these learning paradigms. A key advantage of Hebbian learning is its ability to learn without contrast, much like non-contrastive self-supervised learning techniques, operating purely in an unsupervised manner. Conversely, FF is flexible regarding labels, akin to contrastive self-supervised learning techniques, supporting both unsupervised learning as we demonstrate here with SCFF and supervised learning. This versatility allows FF to be applied across a broader range of tasks and datasets, enhancing its applicability and effectiveness in diverse scenarios.

\textcolor{black}{Another notable distinction between SCFF and SoftHebb lies in the way the classifier is applied to the neural network. SoftHebb attaches a linear classifier to the final layer, utilizing all neurons from the last layer as inputs. In contrast, SCFF applies an additional pooling operation to aggregate information from intermediate layers, reducing the total number of input neurons before feeding them into the linear classifier. This results in a more compact and efficient representation. For example, on CIFAR-10, SoftHebb employs 24,576 input neurons for classification, whereas SCFF only uses 18,432 due to the added pooling operation. Similarly, for STL-10, SoftHebb utilizes 221,184 input neurons, while SCFF significantly reduces this number to 38,400. This reduction in the number of neurons helps simplify the classification process and contributes to computational efficiency.}

\textcolor{black}{Additionally, our method provides a computational advantage. Table \ref{tab:trainingtime} presents a comparison of single-layer training time per epoch on the CIFAR-10 dataset. To ensure a fair comparison, all methods are evaluated under a common setting: a single convolutional layer with $5 \times 5$ filters, 3 input channels, and 96 output channels, using their respective training methodologies. The results show that SCFF enables more efficient training of convolutional layers compared to prior methods. This efficiency stems from SCFF’s learning rule, which eliminates weight transpositions and matrix multiplications during gradient computation, thereby reducing computational complexity. In contrast, algorithms such as SoftHebb rely on these operations, leading to increased training time.}

\subsection{Comparison to energy-based learning methods}

Energy-based learning methods (Table \ref{tab:compareEnergy}), such as Equilibrium Propagation (EP), Dual Propagation (DP) and Latent Equilibrium (LE) \cite{scellier2017equilibrium, hoier2023dual,haider2021latent}, also offer locality in space and time. These methods have a significant advantage over SCFF due to their strong mathematical foundations, closely approximating gradients from BP and backpropagation through time (BPTT). This theoretical rigor allows them to be applied to a wide range of physical systems, making them particularly appealing for neuromorphic computing applications. EP, for instance, can operate in an unsupervised manner \cite{unsuperviseep}, while recent advancements in Genralized Latent Equilibrium (GLE) \cite{ellenberger2024backpropagation} have extended these models to handle sequential data effectively.

However, the implementation of energy-based methods poses certain challenges. Specifically, the backward pass in these methods requires either bidirectional neural networks or dedicated backward circuits \cite{kendall2020training,yi2023activity}. These requirements can be complex to design and build in a manner that is both energy-efficient and compact. In contrast, the simplicity and versatility of SCFF in supporting both supervised and unsupervised learning, without the need for complex backward circuitry, make it a practical alternative for many applications \cite{momeni2024trainingphysicalneuralnetworks}. This balance of accuracy, ease of implementation, and versatility underscores the potential of SCFF in advancing neuromorphic computing and biologically inspired learning systems.

\begin{table}
\centering
\footnotesize
\caption{Test accuracy [\%] comparison of SCFF with energy-based methods on MNIST, CIFAR-10 and STL-10 datasets. The symbols \greencheck, \redcross, and  - mean ``yes", ``no", and  ``no reported results" respectively.}
\label{tab:compareEnergy}
\begin{tabular}{@{}lcccccc@{}}
\toprule
\textbf{Method}  & \textbf{Unsupervised}   & \textbf{MNIST} & \textbf{CIFAR-10} & \textbf{STL-10}  \\ \midrule

EqProp (Laborieux et. al. 2021 \cite{laborieux2021scaling})               & \redcross        & 98.0                           & 88.6         & -    \\
EqProp (Liu et. al. 2024 \cite{unsuperviseep})               & \greencheck        & 97.6                           & 71.5         & -    \\
DualProp (H\o ier et. al. 2023 \cite{hoier2023dual})                   & \redcross   & 98.4                           & 92.3        & -    \\
\textbf{SCFF (ours) }                        & \greencheck             & \textbf{98.7}             & \textbf{80.8}                 & \textbf{77.3}\\
 \bottomrule
\end{tabular}
\end{table}



\section{Conclusion}\label{sec13}

In conclusion, the Forward-Forward (FF) algorithm has sparked significant advancements in both biologically-inspired deep learning and hardware-efficient computation. However, its original form faced challenges in handling complex datasets and time-varying sequential data. Our method, Self Contrastive Forward-Forward (SCFF), addresses these limitations by integrating contrastive self-supervised learning principles directly at the input level, enabling the generation of positive and negative examples though a simple concatenation of input data. SCFF not only surpasses existing unsupervised learning algorithms in accuracy on datasets like MNIST, CIFAR-10, and STL-10 but also successfully extends the FF approach to sequential data, demonstrating its applicability to a broader range of tasks. These developments pave the way for more robust and versatile applications of FF in both neuromorphic computing and beyond, opening new avenues for research and practical implementations in the field.

\section{Methods}\label{secMeth}

SCFF learns representations by maximizing agreement (increasing activations/goodness) between concatenated pairs of identical data examples while minimizing agreement (reducing activations/goodness) between concatenated pairs of different data examples using a cross-entropy-like loss function at each layer. The network is trained layer by layer, with each layer's weights being frozen before moving on to the next. Unlike the original FF framework, this approach incorporates several key components that contribute to achieving high accuracy across various tasks.

\subsection*{Normalization and Standardization}

For vision tasks, the data is typically normalized by subtracting the mean and dividing by the standard deviation for each channel. These mean and standard deviation values are computed across the entire training dataset, separately for each of the three color channels. This dataset-wide normalization centers the data, ensuring that each channel has a mean of 0 and is on a comparable scale.

In addition to dataset-wide normalization, we also applied per-image standardization, which plays an important role in unsupervised feature learning \cite{huang2023normalization}. Standardizing the images involves scaling the pixel values of each image such that the resultant pixel values of the image have a mean of 0 and a standard deviation of 1. This is done before each layer during processing \cite{coates2011analysis, miconi2021hebbian}, ensuring that each sample is centered, which improves learning stability and helps the network handle varying illumination or contrast between images.

\subsection*{Concatenation}

The positive and negative examples (e.g. $\boldsymbol{x}_{i,\text{pos}}^{(0)}$ and $\boldsymbol{x}_{j,\text{neg}}^{(0)}$) are generated by concatenating two identical images for the positive examples and two different images for the negative examples. After being processed by the first layer, the output vectors $\boldsymbol{y}_{i,\text{pos}}^{(0)}$ and $\boldsymbol{y}_{j,\text{neg}}^{(0)}$ are obtained. There are two approaches for generating the inputs to the next layer. The first approach is to directly use the first layer's output of the positive example $\boldsymbol{y}_{i,\text{pos}}^{(0)}$ as the positive input $\boldsymbol{x}_{i,\text{pos}}^{(1)}$, and the first layer output of the negative example $\boldsymbol{y}_{j,\text{neg}}^{(0)}$ as the negative input $\boldsymbol{x}_{j,\text{neg}}^{(1)}$ for the next layer (refer to the highlighted blue section in Algorithm 1 in Appendix \textcolor{black}{D}). The second approach involves re-concatenating to form new positive and negative inputs for the next layer. This is done by treating the first layer's positive outputs as a new dataset and recreating the corresponding positive and negative examples, similar to how the original dataset was processed to generate the initial positive and negative examples (refer to the highlighted blue section in Algorithm 2 in Appendix \textcolor{black}{D}). 

Appendix \textcolor{black}{D} details the workflows of Algorithm 1 and Algorithm 2, focusing on their different approaches to generating positive and negative examples after the first layer. In practice, Algorithm 1 tends to be more effective for training the lower layers immediately following the first layer, while Algorithm 2 shows better performance in training deeper layers. Specifically, for the CIFAR-10 dataset, Algorithm 1 is utilized to train the second layer, while Algorithm 2 is applied to train the third layer. Similarly, for the STL-10 dataset, Algorithm 1 is employed for training the second and third layers, and Algorithm 2 is used for the fourth layer.

\subsection*{Triangle method of transmitting the information}
``Triangle" method was firstly introduced by Coates et al. \cite{coates2011analysis} to compute the activations of the learned features by K-means clustering. This method was later found to be effective in other Hebbian-based algorithms \cite{miconi2021hebbian, journe2022hebbian} for transmitting the information from one layer to the next. The method involves subtracting the mean activation (computed across all channels at a given position) from each channel, and then rectifying any negative values to zero before the pooling layer. This approach to feature mapping can be viewed as a simple form of "competition" between features while also promoting sparsity.

Importantly, the "Triangle" activation only determines the responses passed to the next layer; it does not influence the plasticity. The output used for plasticity at each position is given by $\boldsymbol{y}_{i,\text{pos}}^{(l)} = f^{(l)}(\boldsymbol{x}_{i,\text{pos}})$ and $\boldsymbol{y}_{i,\text{neg}}^{(l)} = f^{(l)}(\boldsymbol{x}_{j,\text{neg}})$, where $f^{(l)}$ refers to the convolutional operations followed by ReLU activation at layer $l$.

\subsection*{Penalty term}

Training with the FF loss can lead to excessively high output activations for positive examples, which significantly drives positive gradients and encourages unchecked growth in their activation. To mitigate this, we introduce a small penalty term—the Frobenius Norm of the Goodness vector—into the training loss function. For outputs from a CNN layer, the goodness vector $G_{i,h,w,\text{pos}}^{(l)}$ is a two-dimensional matrix where each element represents the goodness calculated over the channel outputs processed by all filters under the same receptive field. In the case of Bi-RNN outputs, the goodness vector  $G_{i,t,\text{pos}}^{(l)}$ is a one-dimensional matrix, with each element representing the goodness at each time step. When a large goodness value is computed for a positive example, it generates a negative gradient that reduces the activation, thereby preventing excessive growth. The impact of this penalty term on training performance is further analyzed in Appendix \textcolor{black}{G}.

\subsection*{Additional pooling operation to retrieve the features}

To assess the performance of the intermediate layers in image classification tasks, we apply an additional pooling operation (average or max pooling) to the output of the pooling layer. This reduces the dimensionality of the features and helps in selecting relevant neuron activities. This approach is inspired by the "four quadrant" method used in previous work \cite{coates2011analysis, miconi2021hebbian}, where local regions extracted from the convolutional layer are divided into four equal-sized quadrants, and the sum of neuron activations in each quadrant is computed for downstream linear classification tasks.

Appendix \textcolor{black}{E} provides detailed information on the specific architecture of this additional pooling layer for various tasks.

\textcolor{black}{
\subsection*{Training setup}
All experiments were conducted on a server equipped with an NVIDIA GeForce RTX 4090 GPU (24 GB memory). Training and evaluation were implemented using PyTorch and executed on a single GPU.
}

\backmatter

\bmhead{Data availability}
The datasets used during the current study, i.e.,
IRIS \cite{misc_iris_53}, MNIST \cite{deng2012mnist}, CIFAR-10 \cite{CIFAR-10}, STL-10 \cite{coates2011analysis} and FSDD (Free
Spoken Digit Dataset) \cite{jackson2018jakobovski} , are available online.

\bmhead{Code availability}
The code to reproduce the results is available at: https://github.com/neurophysics-cnrsthales/contrastive-forward-forward

\section*{Declarations}

\bmhead{Acknowledgments}
This work was supported by the European Research Council advanced grant GrenaDyn (reference: 101020684). The text of the article was partially edited by a large language model (OpenAI ChatGPT). The authors would like to thank D. Querlioz for discussion and invaluable feedback.

\bmhead{Author contributions statement}
X.C. and J.G. devised the study. X.C. performed all the simulations and experiments.
X.C., J.G, D. L and J. L actively discussed the results at every stage of the study.
X.C and J.G. wrote the initial version of the manuscript. All authors
reviewed the manuscript.

\bmhead{Competing interests}
The authors declare no competing interests.

\end{document}